\def\year{2019}\relax
\documentclass[letterpaper]{article} 
\usepackage{aaai19}  
\usepackage{times}  
\usepackage{helvet}  
\usepackage{courier}  
\usepackage{url}  
\usepackage{graphicx}  
\usepackage{amsmath}
\usepackage{hhline}
\usepackage{multirow}
\usepackage[multiple]{footmisc}
\newcommand{\argmin}{\mathop{\rm arg~min}\limits}
\begin{document}
%
\title{Unsupervised Cross-lingual Word Embedding \\by Multilingual Neural Language Models}
\author{Takashi Wada\\ Nara Institute of Science and Technology,\\ Nara, Japan \\ wada.takashi.wp7@is.naist.jp  \And Tomoharu Iwata\\ NTT Communication Science Laboratories,\\ Kyoto, Japan\\\ iwata.tomoharu@lab.ntt.co.jp}
\nocopyright
\maketitle
\begin{abstract}
We propose an unsupervised method to obtain cross-lingual embeddings without any parallel data or pre-trained word embeddings. The proposed model, which we call {\it multilingual neural language models}, takes sentences of multiple languages as an input. The proposed model contains bidirectional LSTMs that perform as forward and backward language models, and these networks are shared among all the languages. The other parameters, i.e. word embeddings and linear transformation between hidden states and outputs, are specific to each language. The shared LSTMs can capture the common sentence structure among all languages. Accordingly, word embeddings of each language are mapped into a common latent space, making it possible to measure the similarity of words across multiple languages. We evaluate the quality of the cross-lingual word embeddings on a word alignment task. Our experiments demonstrate that our model can obtain cross-lingual embeddings of much higher quality than existing unsupervised models when only a small amount of monolingual data (i.e. 50k sentences) are available, or the domains of monolingual data are different across languages.
\end{abstract}
\section{Introduction}
Cross-lingual word representation learning has been recognized as a very important research topic in natural language processing (NLP). It aims to represent multilingual word embeddings in a common space, and has been applied to many multilingual tasks, such as machine translation \cite{MT_crossemb} and bilingual named entity recognition \cite{named_entity}. It also enables the transfer of knowledge from one language into another \cite{parse_crossemb,Adams}.

A number of methods have been proposed that obtain multilingual word embeddings. The key idea is to learn a linear transformation that maps word embedding spaces of different languages. Most of them utilize parallel data such as parallel corpus and bilingual dictionaries to learn a mapping \cite{sup_map}. However, such data are not readily available for many language pairs, especially for low-resource languages.

To tackle this problem, a few unsupervised methods have been proposed that obtain cross-lingual word embeddings without any parallel data \cite{MUSE,Zhang,Zhang2,vecmap,vecmap2}. Their methods have opened up the possibility of performing unsupervised neural machine translation \cite{FB_UNMT,UNMT}. \citeauthor{MUSE} \shortcite{MUSE}, \citeauthor{Zhang} \shortcite{Zhang} propose a model based on adversarial training, and similarly \citeauthor{Zhang2} \shortcite{Zhang2} propose a model that employs Wasserstein GAN \cite{WGAN}. Surprisingly, these models have outperformed some supervised methods in their experiments. Recently, however, \citeauthor{limitation} \shortcite {limitation} have pointed out that the model of \citeauthor{MUSE} \shortcite{MUSE} is effective only when the domain of monolingual corpora is the same across languages and languages to align are linguistically similar. \citeauthor{vecmap2} \shortcite{vecmap2}, on the other hand, have overcome this problem and proposed a more robust method that enables to align word embeddings of distant language pairs such as Finnish and English. However, all of these approaches still have a common significant bottleneck: they require a large amount of monolingual corpora to obtain cross-lingual word embedddings, and such data are not readily available among minor languages. 

In this work, we propose a new unsupervised method that can obtain cross-lingual embeddings even in a low-resource setting. We define our method as {\it multilingual neural language model}, that obtains cross-lingual embeddings by capturing a common structure among multiple languages. More specifically, our model employs bidirectional LSTM networks \cite{BiRNN,LSTM} that respectively perform as forward and backward language models \cite{LM}, and these parameters are shared among multiple languages. The shared LSTM networks learn a common structure of multiple languages, and the shared network encodes words of different languages into a common space.  Our model is significantly different from the existing unsupervised methods in that while they aim to align two pre-trained word embedding spaces, ours jointly learns multilingual word embeddings without any pre-training. Our experiments show that our model is more stable than the existing methods under a low-resource condition, where it is difficult to obtain fine-grained monolingual word embeddings. 

\section{Our Model}

\begin{figure}[]
\begin{center}
{\includegraphics[]{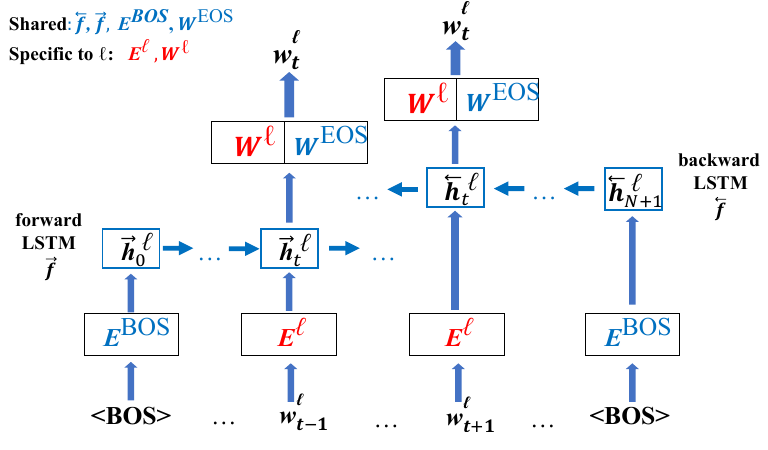}}
\end{center}
\caption{Illustration of our proposed {\it multilingual neural language model}. The parameters shared among across multiple languages are the ones of forward and backward LSTMs $\overrightarrow{f}$ and $\overleftarrow{f}$, the embedding of $<$BOS$>$, $E^{\rm BOS}$, and the linear projection for $<$EOS$>$, $W^{\rm EOS}$. On the other hand, word embeddings, $E^{\ell}$, and linear projection $W^{\ell}$ are specific to each language $\ell$. The shared LSTMs capture a common structure of multiple languages, and that enables us to map word embeddings $E^{\ell}$ of multiple languages into a common space. \label{model}}
\end{figure}
\subsection{Overview}
We propose a model called multi-lingual neural language model, which produces cross-lingual word embeddings in an unsupervised way. Figure \ref{model} briefly illustrates our proposed model. The model consists of the shared parameters among multiple languages and the specific ones to each language. In what follows, we first summerize which parameters are shared or separate across languages:

\begin{itemize}
 \item Shared Parameters
    \begin{itemize}
      \item $\boldsymbol{\overrightarrow{f}}$ and $\boldsymbol{\overleftarrow{f}}$: LSTM networks which perform as forward and backward language models, independently.
      \item $\boldsymbol{{E^{\rm BOS}}}$: The embedding of $<$BOS$>$, an initial input to the language models.
      \item $\boldsymbol{W^{\rm EOS}}$: The linear mapping for $<$EOS$>$, which calculates how likely it is that the next word is the end of a sentence.
     \end{itemize}
 \item Separate Parameters 
    \begin{itemize}
      \item $\boldsymbol{E^{\ell}}$: Word embeddings of language $\ell$
      \item  $\boldsymbol{W^{\ell}}$: Linear projections of language $\ell$, which is used to calculate the probability distribution of the next word.
     \end{itemize}
\end{itemize}

The LSTMs $\overrightarrow{f}$ and $\overleftarrow{f}$ are shared among multiple languages and capture a common language structure. On the other hand, the word embedding function $E^{\ell}$ and liner projection $W^{\ell}$ are specific to each language $\ell$. Since different languages are encoded by the same LSTM functions, similar words across different languages should have a similar representation so that the shared LSTMs can encode them effectively. For instance, suppose our model encodes an English sentence ``He drives a car." and its Spanish translation ``El conduce un coche." In these sentences, each English word corresponds to each Spanish one in the same order. Therefore, these equivalent words would have similar representations so that the shared language models can encode the English and Spanish sentences effectively. Although in general each language has its different grammar rule, the shared language models are trained to roughly capture the common structure such as a common basic word order rule (e.g. subject-verb-object) among different languages. Sharing $<$BOS$>$ and $<$EOS$>$ symbols further helps to obtain cross-lingual representations, ensuring that the beginning and end of the hidden states are in the same space regardless of language. In particular, sharing $<$EOS$>$ symbol indicates that the same linear function predicts how likely it is that the next word is the end of a sentence. In order for the forward and backward language models to predict the end of a sentence with high probability, the words that appear near the end or beginning of a sentence such as punctuation marks and conjunctions should have very close representations among different languages. 

\subsection{Network Structure}
Suppose a sentence with $N$ words in language $\ell$, $\langle w^\ell_{1},w^\ell_{2}, ..., w^\ell_{N} \rangle$. The forward language model calculates the probability of upcoming word $w^\ell_{t}$ given the previous words  $w^\ell_{1},w^\ell_{2}, ..., w^\ell_{t-1}$. 
\begin{equation}
P( w^\ell_{1},w^\ell_{2},..., w^\ell_{N}) = \prod_{t=1}^N p(w^\ell_{t}|w^\ell_{1},w^\ell_{2},..., w^\ell_{t-1}). 
\end{equation}
The backward language model is computed similarly given the backward context:
\begin{equation}
P( w^\ell_{1},w^\ell_{2},..., w^\ell_{N}) = \prod_{t=1}^N p(w^\ell_{t}|w^\ell_{t+1},w^\ell_{t+2},..., w^\ell_{N}). 
\end{equation}
The $t$th hidden states ${h}^\ell_{t}$ of the forward and backward LSTMs are calculated based on the previous hidden state and word embedding,
\begin{equation}
\overrightarrow{h}^\ell_{t} = \overrightarrow{f}(\overrightarrow{h}^\ell_{t-1}, x_{t-1}^\ell),
\end{equation}
\begin{equation}
\overleftarrow{h}^\ell_{t} = \overleftarrow{f}(\overleftarrow{h}^\ell_{t+1}, x_{t+1}^\ell),
\end{equation}
\begin{equation}
x^\ell_{t} =
\begin{cases}
E^{\rm BOS} & \text{if \textit{t} = 0 or \textit{N}+1},\\
E^\ell(w^\ell_{t}) & \text{otherwise}, 
\end{cases}
\end{equation}
where $\overrightarrow{f}(\cdot)$ and $\overleftarrow{f}(\cdot)$ are
the standard LSTM functions. $E^{\rm BOS} $ is the embedding of $<$BOS$>$, which is shared among all the languages. Note that the same word embedding function $E^\ell$ is used among the forward and backward language models.  
The probability distribution of the upcoming word $w^\ell_{t}$  is calculated by the forward and backward models \textbf{independently} based on their current hidden state:
\begin{equation}
p(w^\ell_{t}|w^\ell_{1},w^\ell_{2},..., w^\ell_{t-1}) = {\rm softmax}(g^\ell(\overrightarrow{h}^\ell_{t}))),
\end{equation}
\begin{equation}
p(w^\ell_{t}|w^\ell_{t+1},w^\ell_{t+2},..., w^\ell_{N}).  = {\rm softmax}(g^\ell(\overleftarrow{h}^\ell_{t})), 
\end{equation}
\begin{equation}
g^\ell(h^\ell_{t})  =[W^{\rm EOS}(h^\ell_{t}),W^\ell(h^\ell_{t})],
\end{equation}
where $[x, y]$ means the concatenation of $x$ and $y$. $W^{\rm EOS}$ is a matrix with the size of ($1\times d$), where $d$ is the dimension of the hidden state. This matrix is a mapping function for $<$EOS$>$, and shared among all of the languages. $W^\ell$ is a matrix with the size of ($V^\ell \times d$), where $V^\ell$ is the vocabulary size of language $\ell$ excluding $<$EOS$>$. Therefore, $g$ is a linear transformation with the size of ($(V^\ell+1)\times d$). As with the word embeddings, the same mapping functions are used among the forward and backward language models. 

The largest difference between our model and a standard language model is that our model shares LSTM networks among different languages, and the shared LSTMs capture a common structure of multiple languages. Our model also shares $<$BOS$>$ and $<$EOS$>$ among languages, which encourages word embeddings of multiple languages to be mapped into a common space. 

The proposed model is trained by maximizing the log likelihood of the forward and backward directions for each language $\ell$: 
\begin{equation*}
\begin{split}
\sum_{l=1}^{L}\sum_{i=1}^{S^\ell}\sum_{t=1}^{N^i} &\log p(w^\ell_{i, t}|w^\ell_{i,1},w^\ell_{i, 2},...w^\ell_{i, t-1};\overrightarrow{\theta})\\
& + \log p(w^\ell_{i, t}|w^\ell_{i, t+1},w^\ell_{i, t+2},...w^\ell_{i, N^i}; \overleftarrow{\theta}),
\end{split}
\end{equation*}
where $L$ and $S^\ell$ denote the number of languages and sentences of language $\ell$. $\overrightarrow{\theta}$ and $\overleftarrow{\theta}$ denote the parameters for the forward and backward LSTMs $\overrightarrow{f}$ and $\overleftarrow{f}$, respectively.
\section{Related Work}
\subsection{Unsupervised Word Mapping}
A few unsupervised methods have been proposed that obtain cross-lingual representations in an unsupervised way. Their goal is to find a linear transformation that aligns pre-trained word embeddings of multiple languages. For instance, \citeauthor{vecmap} \shortcite{vecmap} obtain a linear mapping using a parallel vocabulary of automatically aligned digits (i.e. 1-1, 2-2, 15-15...). In fact, their method is weakly supervised because they rely on the aligned information of Arabic numerals across languages. \citeauthor{Zhang} \shortcite{Zhang} and \citeauthor{MUSE} \shortcite{MUSE}, on the other hand, propose fully unsupervised methods that do not make use of any parallel data. Their methods are based on adversarial training \cite{adversarial}: during the training, a discriminator is trained to distinguish between the mapped source and the target embeddings, while the mapping matrix is trained to fool the discriminator. \citeauthor{MUSE} \shortcite{MUSE} further refine the mapping obtained by the adversarial training. They build a synthetic parallel vocabulary using the mapping, and apply a supervised method given the pseudo parallel data. \citeauthor{Zhang2} \shortcite{Zhang2} employ Wasserstein GAN and obtain cross-lingual representations by minimizing the earth-mover's distance. \citeauthor{vecmap2} \shortcite{vecmap2} propose an unsupervised method using a significantly different approach from them. It first roughly aligns words across language using structural similarity of word embedding spaces, and refines the word alignment by repeating a robust self-learning method until convergence. They have found that their approach is much more effective than \citeauthor{Zhang} \shortcite{Zhang} and \citeauthor{MUSE} \shortcite{MUSE} on realistic scenarios, namely when languages to align are linguistically distant or training data are non-comparable across language. 

The common objective among all these unsupervised methods is to map word embeddings of multiple languages into a common space. In their experiments. the word embeddings are pre-trained on a large amount of monolingual data such as Wikipedia before their methods are applied. Therefore, they haven't evaluated their method on the condition when only a small amount of data are available. That condition is very realistic for minor languages, and an unsupervised method can be very useful for these languages. In our experiments, it turns out that existing approaches do not perform well without enough data, while our proposed method can align words with as small data as fifty thousand sentences for each language.

\subsection{Siamese Neural Network}
Our model embeds words of multiple languages into a common space by sharing LSTM parameters among the languages. In general, the model architecture of sharing parameters among different domains is called the ``Siamese Neural Network" \cite{Siamese}. It is known to be very effective at representing data of different domains in a common space, and this technique has been employed in many NLP tasks. For example, \citeauthor{zeroshot_NMT} \shortcite{zeroshot_NMT} built a neural machine translation model whose encoder and decoder parameters are shared among multiple languages. They have observed that sentences of multiple languages are mapped into a common space, and that has made it possible to perform zero-shot translation. \citeauthor{named_entity} \shortcite{named_entity} share LSTM networks of their named entity recognition model across multiple languages, and improve the performance in resource-poor languages. Note that these models are fully supervised and require parallel data to obtain cross-lingual representations. Our model, on the other hand, does not require any parallel or cross-lingual data, and it acquires cross-lingual word embeddings through finding a common language structure in an unsupervised way.
\section{Experiments }

\subsection{Data sets}
We considered two learning scenarios that we deem realistic for low-resource languages: 
\begin{enumerate}
 \item Only a small amount of monolingual data are available. 
 \item The domains of monolingual corpora are different across languages.
\end{enumerate}
For the first case, we used the News Crawl 2012 monolingual corpus for every language except for Finnish, for which we used News Crawl 2014. These data are provided by WMT2013\footnote{\url{http://www.statmt.org/wmt13/translation-task.html}} and 2017\footnote{\url{http://www.statmt.org/wmt17/translation-task.html}}.  We randomly extracted 50k sentences in each language, and used them as training data. We also extracted 100k, 150k, 200k, and 250k sentences and analyzed the impact of the data size. For the second scenario, we used the Europarl corpus \cite{Europarl} as an English monolingual corpus, and the News Crawl corpus for the other languages. We randomly extracted one million sentences from each corpus and used them as training data. The full vocabulary sizes of the Europarl and News Crawl corpora in English were 79258 and 265368 respectively, indicating the large difference of the domains.  We did not use any validation data during the training. We tokenized and lowercased these corpora using Moses toolkit\footnote{\url{https://github.com/moses-smt/mosesDecoder}}. We evaluated models in the pairs of \{French, German, Spanish, Finnish, Russian, Czech\}-English. 
\subsection{Evaluation} 

\begin{table*}[!]  
\hbox to\hsize{\hfil
\makebox[\linewidth]{
\scalebox{0.9}[1.0]{
\centering
\begin{tabular}{|l|r|r||r|r||r|r||r|r||r|r||r|r|}\hline
& \multicolumn{2}{c||}{fr-en} &\multicolumn{2}{c||}{de-en} &\multicolumn{2}{c||}{es-en} &\multicolumn{2}{c||}{fi-en}  &\multicolumn{2}{c||}{ru-en}  &\multicolumn{2}{c|}{cs-en} \\\cline{2-13}
& p@1 &p@5 &p@1  &p@5  &p@1 &p@5  &p@1 &p@5&p@1 &p@5&p@1 &p@5  \\\hline
\textsc{Random}   &0.1&0.5&0.1&0.5&0.1&0.5&0.1&0.5&0.1&0.5&0.1&0.5 \\
\citeauthor{MUSE} \shortcite{MUSE}  &2.5&7.7&0.6&3.5&3.0&9.0 &0.0   &0.4  &0.1   &0.7 &0.0   &1.2\\
\citeauthor{MUSE} \shortcite{MUSE} + normalize  &0.7&3.0&0.6&3.3 &0.5 &2.6 &0.0&0.4 &0.0&0.5 &0.1&0.3 \\
\citeauthor{vecmap2} \shortcite{vecmap2}  &2.4&6.8&1.0&4.5&1.0 &5.0&0.0 &0.1 &0.2 &0.9 &0.4   &1.6 \\
\textsc{Ours}&\textbf{7.3}&\textbf{16.5}&\textbf{4.6}&\textbf{12.0}&\textbf{8.2} &\textbf{18.0}  &\textbf{2.7}   &\textbf{7.3}  &\textbf{2.7}   &\textbf{6.9} &\textbf{3.7}  &\textbf{10.2} \\\hline
\end{tabular}\hfil
}
}
}
\caption{ Word alignment average precisions p@1 and 5 when models are trained on 50k sentences of source and target languages.\label{low_resource}}
\end{table*}

\begin{table*}[!]  
\hbox to\hsize{\hfil
\makebox[\linewidth]{
\scalebox{0.9}[1.0]{
\begin{tabular}{|l|r|r||r|r||r|r||r|r||r|r||r|r|}\hline
& \multicolumn{2}{c||}{fr-en} &\multicolumn{2}{c||}{de-en} &\multicolumn{2}{c||}{es-en} &\multicolumn{2}{c||}{fi-en}  &\multicolumn{2}{c||}{ru-en}  &\multicolumn{2}{c|}{cs-en} \\\cline{2-13}
& p@1 &p@5 &p@1  &p@5  &p@1 &p@5  &p@1 &p@5&p@1 &p@5&p@1 &p@5  \\\hline
\textsc{Random}   &0.1&0.5&0.1&0.5&0.1&0.5&0.1&0.5&0.1&0.5&0.1&0.5 \\
\citeauthor{MUSE} \shortcite{MUSE}  &0.8&4.2&0.2&1.3&1.4&4.6 &0.1   &0.6  &0.6   &2.1 &0.5   &1.3\\
\citeauthor{MUSE} \shortcite{MUSE} + normalize  &0.2 &1.2 &0.1&0.8&0.2 &1.0 &0.2&1.1&0.3&1.1 &0.3 &1.2\\
\citeauthor{vecmap2} \shortcite{vecmap2}  &6.1&14.7&1.1&5.0&\textbf{29.9} &\textbf{45.3}&0.5 &2.2 & 0.1&1.2 &0.5   &2.2  \\
\textsc{Ours}&\textbf{12.7}&\textbf{26.6}&\textbf{3.4}&\textbf{10.0}&14.9 &28.6 &\textbf{3.0}   &\textbf{8.5}  &\textbf{3.8}   &\textbf{11.1} &\textbf{4.0}   &\textbf{10.8}\\\hline
\end{tabular}\hfil
}
}
}
\caption{ Word alignment average precisions p@1 and 5 when models are trained on one million sentences extracted from different domains between source and target languages. \label{diff_domains}}
\end{table*}

\begin{table*}[!t]  
\hbox to\hsize{\hfil
\centering
\begin{tabular}[c]{| l | l | l | l |  | l | l | l | }\hline
\multirow{ 2}{*}{source word (es)} &\multicolumn{3}{c||}{ \textsc{Ours} } &\multicolumn{3}{c|}{\citeauthor{vecmap2} \shortcite{vecmap2}} \\\cline{2-7}
& top 1 & top 2  & top 3  & top 1 & top 2  & top 3   \\\hline
acusado &\textbf{accused} &designed &captured& english & dark & drama \\\hline
actor& \textbf{actor}& artist& candidate & appointment& \textbf{actor} &charlie\\\hline
casi &\textbf{almost}& approximately& about &around &age &capita \\\hline
aunque &\textbf{although}& but &drafting &about & are & been\\\hline
d\'{i}as &\textbf{days} &decades &decade &bodies & both &along \\\hline
actualmente& \textbf{currently} &clearly& essentially &comes &continued &candidates\\\hline
contiene &\textbf{contains} &defines& constitutes &barrier &etiquette& commissioned\\\hline
cap\'{i}tulo &\textbf{chapter} &episode& cause &arriving &bulls& dawn\\\hline

\end{tabular}\hfil
}
\caption{Some examples when Spanish and English words matched correctly by our model using 50k sentences, but not by \citeauthor{vecmap2} \shortcite{vecmap2}. Each column indicates 1st, 2nd, and 3rd most similar English words to each Spanish word. English words in bold font are translations of each Spanish word.\label{words}}
\end{table*}

In this work, we evaluate our methods on a word alignment task. Given a list of $M$ words in a source language s $[x_{1},x_{2},..., x_{M}]$ and target language $t$ $[y_{1},y_{2},..., y_{M}]$, the word alignment task is to find one-to-one correspondence between these words. If a model generates accurate cross-lingual word embeddings, it is possible to align words properly by measuring the similarity of the embeddings. In our experiment, we used the bilingual dictionary data published by \citeauthor{MUSE} \shortcite{MUSE}, and extracted 1,000 unique pairs of words that are included in the vocabulary of the News Crawl data of from 50k to 300k sentences. As a measurement of the word embeddings, we used cross-domain similarity local scaling (CSLS), which is also used in \citeauthor{MUSE} \shortcite{MUSE} and \citeauthor{vecmap2} \shortcite{vecmap2} . CSLS can mitigate the hubness problem in high-dimensional spaces, and can generally improve matching accuracy. It takes into account the mean similarity of a source language embedding $x$ to its $K$ nearest neighbors in a target language:

\begin{equation}
rT(x)=\frac{1}{K} \sum_{y \in {\mathcal N_{T}(x)}} {\rm cos} (x,y),
\end{equation}
where cos is the cosine similarity and $N_{T}(x)$ denotes the $K$ closest target embeddings to $x$. Following their suggestion, we set $K$ as 10. $rR(y)$ is defined in a similar way for any target language embedding y. CSLS(x, y) is then calculated as follows:
\begin{equation}
{\rm CSLS}(x, y) = 2 {\rm cos}(x, y) -rT(x)- rS(y).
\end{equation}

For each source word $x_{i}$, we extracted the $k$ target words that have the highest CSLS scores ($k$ = 1 or 5). However, since the value of $rT(x)$ does not affect the result of this evaluation, we omit the score from CSLS in our experiments. We report the precision p@k: how often the correct translation of a source word $x_{i}$ is included in the $k$ extracted target words.
\subsection{Baseline} 
As baselines, we compared our model to that of \citeauthor{MUSE} \shortcite{MUSE} and \citeauthor{vecmap2} \shortcite{vecmap2}. \citeauthor{MUSE} \shortcite{MUSE} aim to find a mapping matrix $W$ based on adversarial training. The discriminator is trained to distinguish the domains (i.e. language) of the embeddings, while the mapping is trained to fool the discriminator. Then, $W$ is used to match frequent source and target words, and induce a bilingual dictionary. Given the pseudo dictionary, a new mapping matrix $W$ is then trained in the same manner as a supervised method, which solves the Orthogonal Procrustes problem:

\begin{equation*}
\begin{split}
&W^* = \argmin_{W}  \|WX-Y\|_{F} = UV^{\mathrm{T}},\\ & s.t. ~~~ U\sum{}V^{\mathrm{T}} = {\rm SVD}(YX^{\mathrm{T}}).
\end{split}
\end{equation*}
This training can be iterated using the new matrix $W$ to induce a new bilingual dictionary. This method assumes that the frequent words can serve as reliable anchors to learn a mapping. Since they suggest normalizing word embeddings in some language pairs, we evaluated their method with and without normalization. \citeauthor{vecmap2} \shortcite{vecmap2} use a different approach and employ a robust self-learning method. First, they roughly align words based on the similarity of word emebeddings. Then, they repeat the self-learning approach, where they alternatively update a mapping function and word alignment.

To implement the baseline methods, we used the code published by the authors\footnote{\url{https://github.com/facebookresearch/MUSE}}\footnote{\url{https://github.com/artetxem/vecmap}}. To obtain monolingual word embeddings, we used word2vec \cite{word2vec}. Note that these embeddings were used only for the baselines, but not for ours since our method does not require any pre-trained embeddings. For a fair comparison, we used the same monolingual corpus with the same vocabulary size for the baselines and our model. 
\subsection{Training Settings } 

We preprocessed monolingual data and generated mini-batches for each language. For each iteration, our model alternately read mini-batches of each language, and updated its parameters every time it read one mini-batch. We trained our model for 10 epochs with the mini-batch size 64. The size of word embedding was set as 300, and the size of LSTM hidden states was also set as 300 for the forward and backward LSTMs, respectively. Dropout \cite{dropout} is applied to the hidden state with its rate 0.3. We used SGD \cite{SGD} as an optimizer with the learning rate 1.0. Our parameters, which include word embeddings, were uniformly initialized in [-0.1, 0.1], and gradient clipping \cite{clipping} was used with the clipping value 5.0. We included in the vocabulary the words that were used at least a certain number of times. For the News Crawl corpus, we set the threshold as 3, 5, 5, 5 ,5, 10, and 20 for 50k, 100k, 150k, 200k, 250k, 300k and 1m sentences. For the Europarl corpus, we set the value as 10. We fed 10000 frequent words into the discriminator in \citeauthor{MUSE} \shortcite{MUSE}.

As a model selection criterion, we employed a similar strategy used in the baseline. More specifically, we considered the 3,000 most frequent source words, and used CSLS excluding $rT(x)$ to generate a translation for each of them in a target language. We then computed the average CSLS scores between these deemed translations, and used them as a validation metric.
\section{Results} 

\begin{figure}[!h] 
\begin{center}
{\includegraphics[]{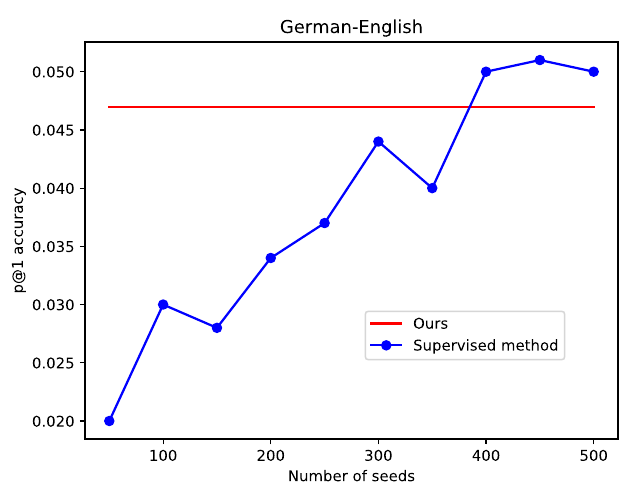}}
\end{center}
\caption{Comparison of p@1 accuracy of German-English pair between supervised word mapping method and our model on 50k sentences. The x axis indicates the number of pairs of words $n$ (= 0,50,100,150,..., 450, 500) that were used for the supervised method, but not for ours, to map word embedding spaces in two languages.  \label{vs_supervised}}
\end{figure}
\begin{figure}[h!] 
\begin{center}
\includegraphics[]{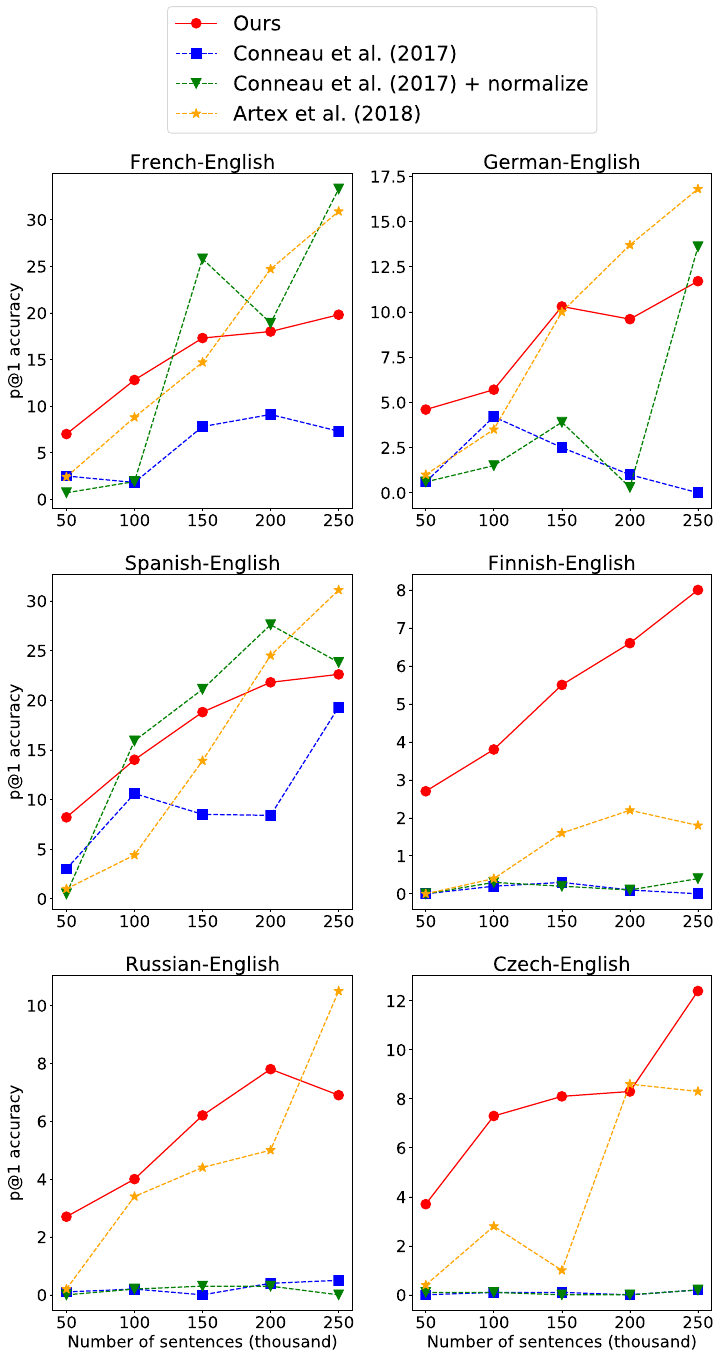}
\end{center}
\caption{Graphs show the change in p@1 accuracy of each language pair as the size of training data increases. The x-axis denotes the number of sentences (thousand) in the monolingual training data of the source and target languages. \label{change_N}}
\end{figure}

\subsection{Bilingual Word Embeddings}
First, we trained our model and obtained cross-lingual embeddings between two languages for each language pair. We report our results under the two scenarios that we considerted realistic when dealing with minor languages. In the first scenario, we trained our model on a very small amount of data, and in the second scenario the model was trained on a large amount of data extracted from different domains between source and target languages. 

Table \ref{low_resource} illustrates the results of the word alignment task under the low-resource scenario. \textsc{Random} is the expected accuracy when words are aligned at random. The result shows that our model outperformed the baseline methods significantly in all of the language pairs, indicating that ours is more robust in a low-resource senario. On the other hand, the baseline methods got poor performance, especially in the Finnish and English pair. Even though \citeauthor{vecmap2} \shortcite{vecmap2} report that their method achieves good performance in that language pair, our experiment has demonstrated that it does not perform well without a large amount of data.

 Table \ref{diff_domains} shows the results when the domains of training data used to obtain source and target embeddings are different. Our method again outperformed the baselines to a large extent except for the Spanish-English pair. The poor performance of \citeauthor{MUSE} \shortcite{MUSE} in such a setting has also been observed in \citeauthor{limitation} \shortcite{limitation}, even though much larger data including Wikipedia were used for training in their experiments. 
 
 Table \ref{words} shows some examples when Spanish and English words were correctly matched by our model, but not by \citeauthor{vecmap2} \shortcite{vecmap2} under the low-resource scenario. 
 The table lists the three most similar English words to each Spanish source word. Our method successfully matched similar or semantically related words to the source words, indicating that our method obtained good cross-lingual embeddings. For example, to the Spanish source word ``casi", our model aligned its translation ``almost" and also very similar words ``approximately" and ``about". Indeed, many of the aligned target words in our model have the same part of speech tag as that of the source word, suggesting that our model captured a common language structure such as rules of word order and roles of vocabulary by sharing LSTMs. On the other hand, \citeauthor{vecmap2} \shortcite{vecmap2} could not align words properly, and there do not seem to exist consistent relations between the source and extracted words. 

\begin{table*}[]  
\centering
\hbox to\hsize{\hfil
\begin{tabular}{|l|r|r|r||r|r|r||r|r|r|}\hline
& \multicolumn{3}{c||}{fr-en} &\multicolumn{3}{c||}{de-en} &\multicolumn{3}{c|}{es-en} \\\cline{2-10}
&50k &100k&300k  &50k &100k&300k  &50k &100k&300k  \\\hline
\textsc{\textsc{Ours (bilingual)} }&{7.3}&\textbf{12.8}&{23.1}&{4.6}&{5.7}&{13.1}&\textbf{8.2} &\textbf{14.0} &  \textbf{27.9}  \\\hline
\textsc{\textsc{Ours (quadrilingual)} }&\textbf{8.4}&{12.3}&\textbf{25.6}&\textbf{4.7}&\textbf{8.4}&\textbf{16.6}&{7.6} &{13.8} & {25.3}  \\\hline
\end{tabular}\hfil
}
\caption{Word alignment average precisions p@1 in each language pair when 50k, 100k, and 300k sentences were used for training. \textsc{Ours (bilingual)}  denotes the accuracy of the models that read source and target languages, generating bilingual word embeddings. \textsc{Ours (quadrilingual)} denotes the accuracy of one model that reads all four languages, producing quadrilingual word embeddings. \label{quad}}
\end{table*}

\subsubsection{Comparison to Supervised Method}
To further investigate the effectiveness of our model, we compared our method to a supervised method under the low-resource setting. The method is a slight modification of \citeauthor{MUSE} \shortcite{MUSE}: it is trained using a bilingual dictionary, and learns a mapping from the source to the target space using iterative Procrustes alignment. We used the code provided by \citeauthor{MUSE} \shortcite{MUSE}. Figure \ref{vs_supervised} compares p@1 accuracy between the supervised method and our model in the German-English pair. The x-axis denotes the number of seeds of the bilingual dictionary that were used for the supervised method, but not for ours. The figure illustrates that our method achieved a better result than the supervised method when the number of seeds was less than 400, which is surprising given that our model is fully unsupervised. 

\subsubsection{Impact of Data Size}
We changed the size of the training data by 50k sentences, and analyzed how the performance of the baselines and our model changed. Figure \ref{change_N} illustrates how the performance changed depending on the data size. It shows that our method achieved a comparable or better result than the baseline methods in all of the language pairs when the number of sentences was not more than 100k. In the closely related language pairs such as \{French, German, Spanish\}-English, the baselines performed better when there were enough amount of data. Among the distant languages such as \{Finnish, Czech\}-English, our model achieved better results overall, while the baseline methods, especially \citeauthor{MUSE} \shortcite{MUSE} got very poor results.

\begin{table}[]
\centering
\begin{tabular}{|l|l|l|l|}
\hline
En (source)  & Fr top1 & De top1 & Es top1 \\ \hline
declared & d\'{e}clar\'{e}    & erkl\"{a}rt    & declarado   \\ \hline
always   & toujours   & immer      & siempre     \\ \hline
are      & sont       & sind       & est\'{a}n       \\ \hline
after    & apr\`{e}s      & nachdem    & despu\'{e}s     \\ \hline
died     & d\'{e}c\'{e}d\'{e}     & starb      & murieron    \\ \hline
\end{tabular}
\caption{Examples of words that were correctly aligned by \textsc{Ours (quadrilingual)} among the four languages.  \label{quadalign}}
\end{table}

\subsection{Quadrilingual Word Embeddings}

Our results of the word alignment task have shown that our model can jointly learn bilingual word embeddings by capturing the common structure of two languages. This success has raised another question: ``Is it also possible to learn a common structure of more than two languages?" To examine this intriguing question, we trained our model that encoded four linguistically similar languages, namely English, French, Spanish, and German,  and aimed to capture the common structure among them. We expect that word embeddings of the four languages should be mapped into a common space, generating what we call quadrilingual word embeddings. Table \ref{quad} describes the result of the word alignment when using bilingual and quadrilingual word embeddings of our model. While quadrilingual word embeddings performed slightly worse than bilingual ones in the Spanish-English pair, they brought large gains in the German-English alignment task and achieved comparable performance overall. Our model successfully mapped word embeddings of the four languages into a common space, making it possible to measure the similarity of words across the multiple languages.

To investigate whether quadrilingual embeddings were actually mapped into a common space, we aligned each English word to French, German and Spanish words in the bilingual dictionary. Table \ref{quadalign} describes the words that were correctly aligned among the four languages. This result indicates that these equivalent words have very similar representations, and that means our model successfully embedded these languages into a common space. Figure \ref{quad} illustrates the scatter plot of the embeddings of the most 1,000 frequent words in each corpus of the four languages. It clearly shows that the word embeddings were clustered based on their meanings rather than their language. For example, the prepositions of the four languages (e.g. de (fr, es), of (en), von (de)) were mapped into the bottom-right area, and determiners (e.g. la (fr, es), the (en), der, die, das (de)) were in the bottom-left area. Near the area where the embedding of `$<$BOS$>$' was mapped, the words from which a new sentence often starts (`,' , et (fr), y(es), und (de), and (en)) were mapped.  
 
\begin{figure}[!h]
\begin{center}
\centering
{\includegraphics[]{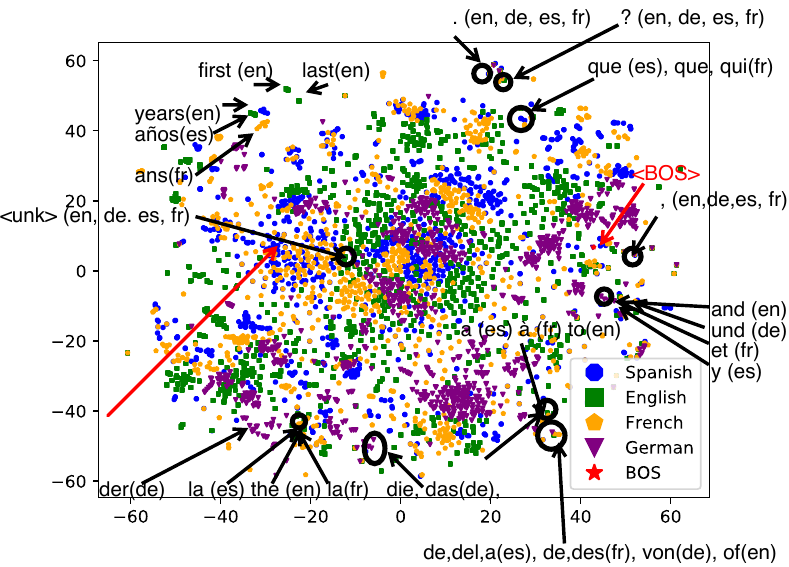}}
\end{center}
\caption{Scatter plot of cross-lingual word embeddings of French, English, German and Spanish obtained by our model. The embeddings are reduced to 2D using tSNE \cite{tSNE}. \label{quad}}
\end{figure}

\section{Conclusion} 
In this paper, we proposed a new unsupervised method that learns cross-lingual embeddings without any parallel data. Our experiments of a word alignment task in six language pairs have demonstrated that our model significantly outperforms existing unsupervised word translation models in all the language pairs under a low resource situation. Our model also achieved better results in five language pairs when the domains of monolingual data are different across language. We also compared our unsupervised method to a supervised one in the German-English word alignment task, and our model achieved a better result than the supervised method that were trained with 350 pairs of words from a bilingual dictionary. Our model also succeeded in obtaining cross-lingual embeddings across four languages, which we call quadrilingual embeddings. These embeddings enabled us to align equivalent words among four languages in an unsupervised way. The visualization of the quadrilingual embeddings showed that these embeddings were actually mapped into a common space, and words with similar meanings had close representations across different languages.
 
Potential future work includes extending our approach to a semi-supervised method that utilizes a bilingual dictionary. One possible idea is to set an additional loss function in our model that decreases the distance of embeddings of equivalent words across languages.

\bibliography{ref}
\bibliographystyle{aaai}
\end{document}